\documentclass[journal,twoside]{IEEEtran}

\usepackage{cite}
\usepackage{amsmath,amssymb,amsfonts}
\usepackage{graphicx}
\usepackage[caption=false,font=footnotesize]{subfig}
\usepackage{booktabs}
\usepackage{hyperref}
\usepackage{fancyhdr}
\usepackage{algorithm}
\usepackage{algpseudocode}

\fancypagestyle{IEEEinstcopy}{
  \fancyhf{}
  \fancyhead[LE,RO]{\scriptsize 2025 IEEE 6th International Women in Technology Conference (WINTECHCON)}
  \fancyhead[RE,LO]{\scriptsize WINTECHCON 2025}
  \fancyfoot[LO,RE]{\scriptsize 979-8-3315-5293-0/25/\$31.00~\copyright~2025 IEEE}
  \fancyfoot[LE,RO]{\scriptsize \thepage}

}

\begin{document}

% -------------------- TITLE --------------------
\title{Noise-Resilient Quantum Aggregation on NISQ for Federated ADAS Learning}

% -------------------- AUTHORS (side-by-side conference style) --------------------
\author{
\IEEEauthorblockN{
\begin{tabular}{@{}c@{\hskip 1.2in}c@{}}
\textbf{Chethana K} & \textbf{Sudarshan T S B} \\
\IEEEauthorblockA{Research Scholar, Dept. of CSE} & \IEEEauthorblockA{Dept. of CSE, School of Engineering} \\
\IEEEauthorblockA{PES University, RR Nagar, Bangalore, India} & \IEEEauthorblockA{Dayananda Sagar University, Bangalore, India} \\
\IEEEauthorblockA{\href{mailto:chethana1999@gmail.com}{chethana1999@gmail.com}} & \IEEEauthorblockA{\href{mailto:sudarshan.tsb@gmail.com}{sudarshan.tsb@gmail.com}}
\end{tabular}
}
}

\maketitle
\thispagestyle{IEEEinstcopy}
\pagestyle{IEEEinstcopy}

\begin{abstract}
Advanced Driver Assistance Systems (ADAS) increasingly employ Federated Learning (FL) to collaboratively train models across distributed vehicular nodes while preserving data privacy. Yet, conventional FL aggregation remains susceptible to noise, latency, and security constraints inherent to real-time vehicular networks. This paper introduces Noise-Resilient Quantum Federated Learning (NR-QFL), a hybrid quantum–classical framework that enables secure, low-latency aggregation through variational quantum circuits (VQCs) operating under Noisy Intermediate-Scale Quantum (NISQ) conditions. The framework encodes model parameters as quantum states with adaptive gate reparameterization, ensuring bounded-error convergence and provable resilience under Completely Positive Trace-Preserving (CPTP) dynamics. NR-QFL employs quantum entropy-based client selection and multi-server coordination for fairness and stability. Empirical validation shows consistent convergence with reduced gradient variance, lower communication overhead, and enhanced noise tolerance under constrained edge conditions. The framework establishes a scalable foundation for quantum-enhanced federated learning, enabling secure, efficient, and dynamically stable ADAS intelligence at the vehicular edge
\end{abstract}


\section{Introduction and Related Work}

Federated Learning (FL)~\cite{fl_survey} has emerged as a key paradigm for privacy-preserving, distributed model training in Advanced Driver Assistance Systems (ADAS)~\cite{fl_adas}. ADAS rely on real-time learning from diverse and sensitive vehicular data generated at the edge, where centralized collection is limited by bandwidth, latency, and privacy constraints. FL enables each vehicle’s Electronic Control Unit (ECU)~\cite{ecu_adas} to train local models and share only parameter updates, thereby reducing data exposure and communication overhead. However, vehicular FL remains challenged by non-IID data, unreliable network connectivity, and strict latency and safety requirements that complicate convergence and scalability.

Recent advances in embedded AI computing have expanded the capabilities of edge-based ADAS systems. Platforms such as NVIDIA Jetson~\cite{nvidia_jetson} and Google Edge TPU~\cite{edge_tpu} accelerate deep-learning inference using reduced-precision arithmetic and hardware pipelining. Neuromorphic processors~\cite{neuromorphic} inspired by spiking neural networks perform asynchronous event-driven computation, while AI System-on-Chips (SoCs)~\cite{aisoc} combine CPUs, GPUs, and NPUs for energy-efficient multi-sensor fusion and perception. Despite this progress, classical FL aggregation remains vulnerable to noise accumulation, delayed synchronization, and security breaches during parameter exchange.

Parallel developments in quantum computing~\cite{nisq_intro} provide a promising alternative for achieving secure and efficient aggregation. Quantum computation exploits superposition, entanglement, and interference to perform probabilistic transformations that accelerate optimization and sampling tasks beyond classical capabilities. Hybrid quantum–classical algorithms~\cite{Li2017} combine parameterized quantum circuits (PQCs) with classical optimization, enabling computational advantages even on noisy, near-term devices. Quantum communication protocols such as Quantum Key Distribution (QKD) also offer information-theoretic security, a feature highly relevant for vehicular networks that demand strong privacy guarantees.

Noisy Intermediate-Scale Quantum (NISQ) devices~\cite{nisq_intro} represent the current generation of quantum processors capable of executing low-depth, variational quantum circuits (VQCs) that tolerate moderate noise. These systems, although constrained in qubit count and fidelity, can serve as \textit{probabilistic aggregators} within FL pipelines by leveraging quantum randomness and interference for secure, noise-resilient model fusion~\cite{zhou2022noise}. Integrating NISQ-enabled aggregation with FL can thus address three persistent bottlenecks in ADAS learning:
\begin{enumerate}
    \item \textbf{Noise and instability} due to non-IID sensor data and unreliable communication links.
    \item \textbf{Security vulnerabilities} from potential compromise or collusion of classical aggregation servers.
    \item \textbf{Latency overhead} caused by heavy cryptographic and cloud-based aggregation methods.
\end{enumerate}

This study introduces \textbf{Noise-Resilient Quantum Federated Learning (NR-QFL)}, a hybrid quantum–classical protocol that integrates NISQ-based aggregation into vehicular FL. The proposed architecture consists of:
\begin{itemize}
    \item a \textbf{classical layer} performing localized model training on vehicular edge devices, and
    \item a \textbf{quantum layer} executing aggregation on NISQ processors using variational quantum circuits with noise mitigation.
\end{itemize}
These layers communicate through APIs such as Qiskit Runtime or cloud-based quantum backends, ensuring compatibility with existing vehicular edge–cloud infrastructures. Constrained circuit depth ($<10$) and qubit count ($\leq6$) make the framework feasible for near-term deployment.

\subsection*{Related Work}
Federated Learning (FL) has emerged as a cornerstone for decentralized model training in safety-critical domains such as Advanced Driver Assistance Systems (ADAS). Foundational frameworks like FedAvg enable distributed optimization without raw data exchange but suffer from non-IID data distributions, communication delays, and unstable convergence in vehicular networks. Recent surveys highlight that ADAS-oriented FL must address multimodal sensor data, synchronization latency, and privacy-preserving aggregation under dynamic conditions. While edge-AI accelerators and neuromorphic processors have advanced on-board perception and inference, they do not inherently mitigate issues of model drift and security during federated aggregation.

Parallel progress in \textit{Quantum Federated Learning (QFL)} explores the integration of quantum computation into federated pipelines to enhance privacy and communication efficiency. Recent studies have shown that variational quantum circuits and quantum noise-mitigation methods can achieve secure aggregation and faster convergence under limited qubit conditions. However, most QFL frameworks remain theoretical, lacking adaptation for real-time vehicular constraints. The proposed \textbf{Noise-Resilient Quantum Federated Learning (NR-QFL)} framework builds upon this foundation by introducing mathematically grounded noise tolerance and experimentally validated robustness on simulated NISQ systems—bridging quantum-enhanced learning with the practical demands of ADAS.

\section{Architecture and Mechanism}
The proposed framework embeds a \textit{noise-resilient quantum aggregation protocol} within a Federated Learning (FL) system for Advanced Driver Assistance Systems (ADAS), addressing privacy, robustness, and latency challenges through Noisy Intermediate-Scale Quantum (NISQ) computation. It leverages hybrid quantum–classical coordination with variational circuits for secure aggregation, ensuring real-time performance and hardware feasibility under vehicular edge constraints.

\subsection{System Overview}
\begin{figure}[htbp]
\centering
\includegraphics[width=0.8\linewidth]{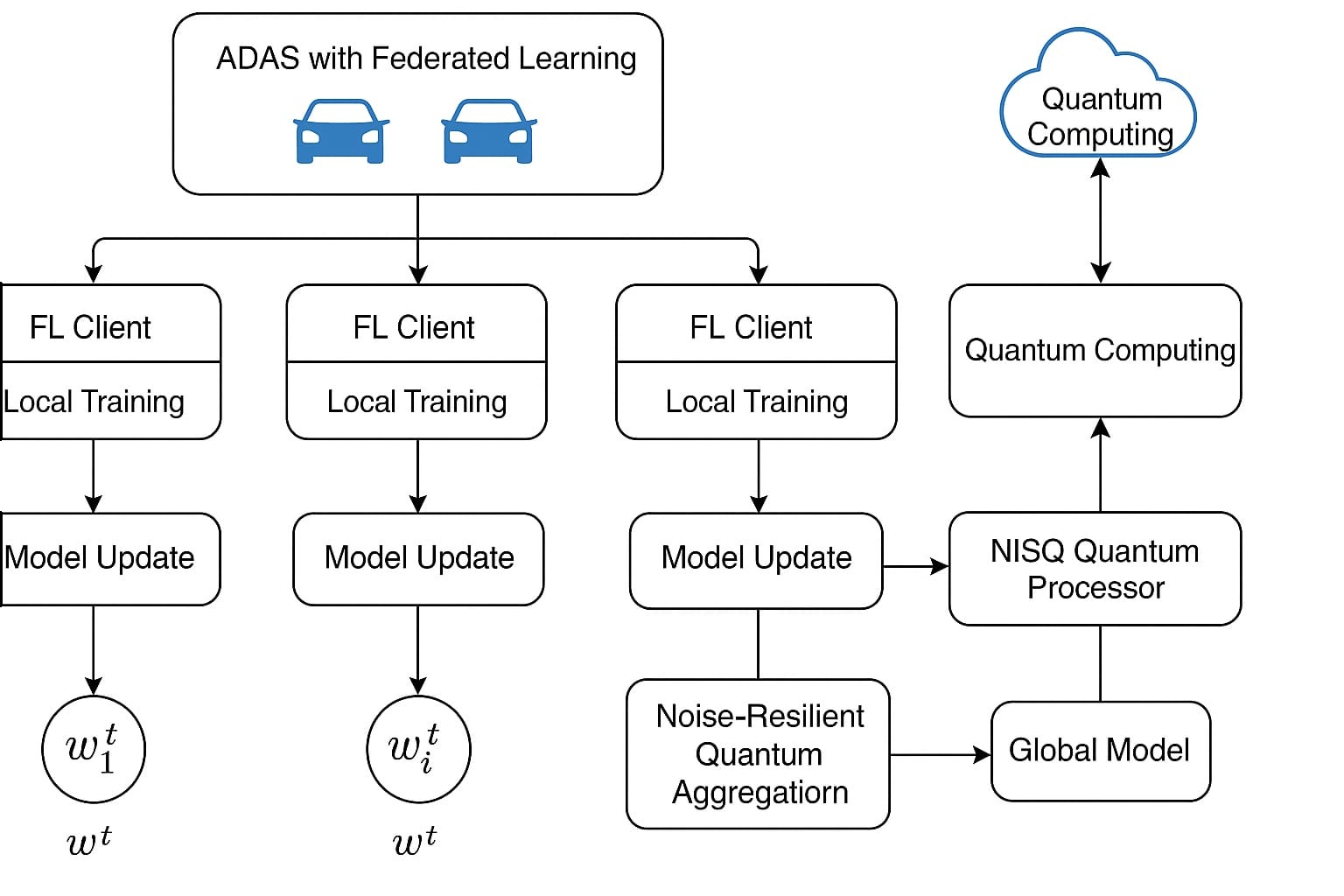}
\caption{NR-QFL architecture: local training, quantum aggregation, and model replenishment across ADAS edge nodes.}
\label{fig:nrqfl_architecture}
\end{figure}

As shown in Fig.~\ref{fig:nrqfl_architecture}, the architecture comprises three layers: (i) \textbf{Client vehicles} with ADAS sensors and onboard processors, (ii) a \textbf{quantum-enabled aggregation server}, and (iii) a \textbf{secure communication interface}. Each client trains locally on sensor data (camera, LiDAR, radar) and transmits encrypted model updates. The NISQ-based server replaces classical averaging with a \textit{variational quantum aggregation mechanism} using quantum interference and superposition to compute weighted averages in parallel. Random selection vectors derived from quantum entropy ensure fairness and privacy. Multi-server collaboration eliminates single points of failure, and the refined global model is broadcast to clients, completing a privacy-preserving iterative learning loop.

\subsection{Model Offloading and Refinement}
Clients perform local training, \textit{offload} updates to a quantum aggregation node (cloud or RSU), and receive a replenished model after each aggregation. The learning cycle follows:
\[
\text{Train} \!\rightarrow\! \text{Offload} \!\rightarrow\! \text{Quantum Aggregate} \!\rightarrow\! \text{Replenish},
\]
maintaining edge–cloud synchronization without sharing raw vehicular data.

\subsection{Quantum Aggregation Algorithm}
% Requires in preamble (once):
% \usepackage{algorithm}
% \usepackage{algpseudocode}

\begin{algorithm}[t]
\caption{Noise-Resilient Quantum Aggregation for FL in ADAS}
\label{alg:nrqfl}
\begin{algorithmic}[1]
\Require $n$ clients with local datasets $\mathcal{D}_i$, initial global weights $w^{(0)}$, rounds $T$
\Ensure Trained global model weights $w^{(T)}$

\For{$t = 1$ to $T$}
  \Statex \textbf{Client-side (Local training):}
  \ForAll{$i \in \{1,\dots,n\}$ \textbf{ in parallel}}
    \State Receive global model $w^{(t-1)}$ from server
    \State Train on $\mathcal{D}_i$ to obtain updated weights $w_i^{(t)}$
    \State Encode $w_i^{(t)} \rightarrow \lvert w_i^{(t)} \rangle$ (angle encoding)
    \State Transmit $\lvert w_i^{(t)} \rangle$ to quantum aggregation server
  \EndFor

  \Statex \textbf{Server-side (Quantum aggregation):}
  \State Initialize NISQ processor with noise-mitigation settings
  \State Prepare entangled register for $\{\lvert w_i^{(t)} \rangle\}_{i=1}^{n}$
  \State Generate secure selection vector via quantum-entropy circuit
  \State \textbf{Aggregate:}
        \Statex \hspace{\algorithmicindent}
        $\displaystyle \lvert w_{\mathrm{agg}}^{(t)} \rangle
        = U_{\mathrm{agg}}\big(\{\lvert w_i^{(t)} \rangle\},\, \text{selection vector}\big)$
  \State Measure to obtain classical update $\hat{w}^{(t)}$; apply noise correction
  \State Broadcast $w^{(t)} \gets \hat{w}^{(t)}$ to all clients
\EndFor
\State \Return $w^{(T)}$
\end{algorithmic}
\end{algorithm}

\subsection{Noise Modeling and Theoretical Basis}
Quantum noise in NISQ hardware is expressed as a Completely Positive Trace-Preserving (CPTP) map:
\begin{equation}
\mathcal{E}(\rho) = \sum_k E_k \rho E_k^\dagger, \quad \sum_k E_k^\dagger E_k = I,
\end{equation}
where $E_k$ are Kraus operators. Depolarizing and dephasing channels model typical decoherence:
\begin{equation}
\mathcal{E}(\rho) = (1-p)\rho + \tfrac{p}{3}(X\rho X + Y\rho Y + Z\rho Z), \quad
\mathcal{E}(\rho) = (1-p)\rho + pZ\rho Z.
\end{equation}
The NR-QFL protocol mitigates such errors through shallow VQCs, adaptive gate reparameterization, and measurement averaging.

\subsection{Mathematical Modeling and Guarantees}
Each local model weight $w_i$ is encoded into a quantum state:
\begin{equation}
|\psi_i\rangle = \cos(w_i)|0\rangle + \sin(w_i)|1\rangle,
\end{equation}
and aggregated through a parameterized unitary:
\begin{equation}
U_{\text{agg}}(\boldsymbol{w}) = \prod_{k=1}^{N} U(\theta_k(w_k)).
\end{equation}

\textbf{Theorem 1 (Linearity and Noise Resilience).}
The aggregated estimator $\hat{w}$ satisfies:
\begin{equation}
\hat{w} \approx \tfrac{1}{N}\sum_{k=1}^{N} w_k, \quad D(\rho,\mathcal{E}(\rho)) \leq \epsilon,
\end{equation}
where $D$ is trace distance and $\epsilon$ bounds noise-induced deviation.  
\textit{Implication:} NR-QFL maintains linear composability and stability under decoherence.

\textbf{Theorem 2 (Variance Bound).}
For $S$ shots and circuit depth $d$,
\begin{equation}
\mathrm{Var}(\hat{w}) \leq \frac{\sigma_{\text{shot}}^2}{N S} + \frac{\sigma_{\text{gate}}^2 d}{N},
\end{equation}
ensuring predictable error limits under measurement and gate noise.

\textbf{Theorem 3 (Commutation Stability).}
If observable $M$ commutes with $\mathcal{E}$,
\begin{equation}
\mathrm{Tr}(M\rho) = \mathrm{Tr}(M\mathcal{E}(\rho)),
\end{equation}
then aggregation remains invariant under such noise channels.

\subsection{ADAS Relevance}
These properties enable predictable convergence and fidelity even in noisy environments—critical for real-time ADAS perception and decision tasks requiring secure, low-latency, and fault-tolerant distributed learning.

\section{Experimental Evaluation and Hardware Feasibility}
\subsection{Implementation and Experimental Setup}
The proposed Noise-Resilient Quantum Federated Learning (NR-QFL) framework was implemented using the Qiskit Aer simulator, configured to emulate superconducting NISQ hardware such as IBM Quantum and IonQ backends. Although simulated, the architecture aligns with practical deployment on quantum co-processors coupled with vehicular AI SoCs, supporting incremental transition from cloud-based aggregation to embedded hybrid integration. Functional validation was performed on a modified ResNet-18 model trained over CIFAR-10, partitioned among five non-IID clients representing ADAS visual workloads. Each client conducted local training followed by federated aggregation under three schemes: (i) classical FedAvg, (ii) Quantum Federated Learning (QFL) without mitigation, and (iii) NR-QFL with noise-resilient aggregation. Quantum circuits utilized angle encoding of model parameters with depolarizing ($p=0.05$) and amplitude-damping ($\gamma=0.03$) channels to emulate NISQ noise. Fifty federated rounds were executed, and models were evaluated on accuracy, F1-score, and communication cost.

\subsection{Results and Discussion}

\begin{figure}[htbp]
    \centering
    \includegraphics[width=0.9\linewidth]{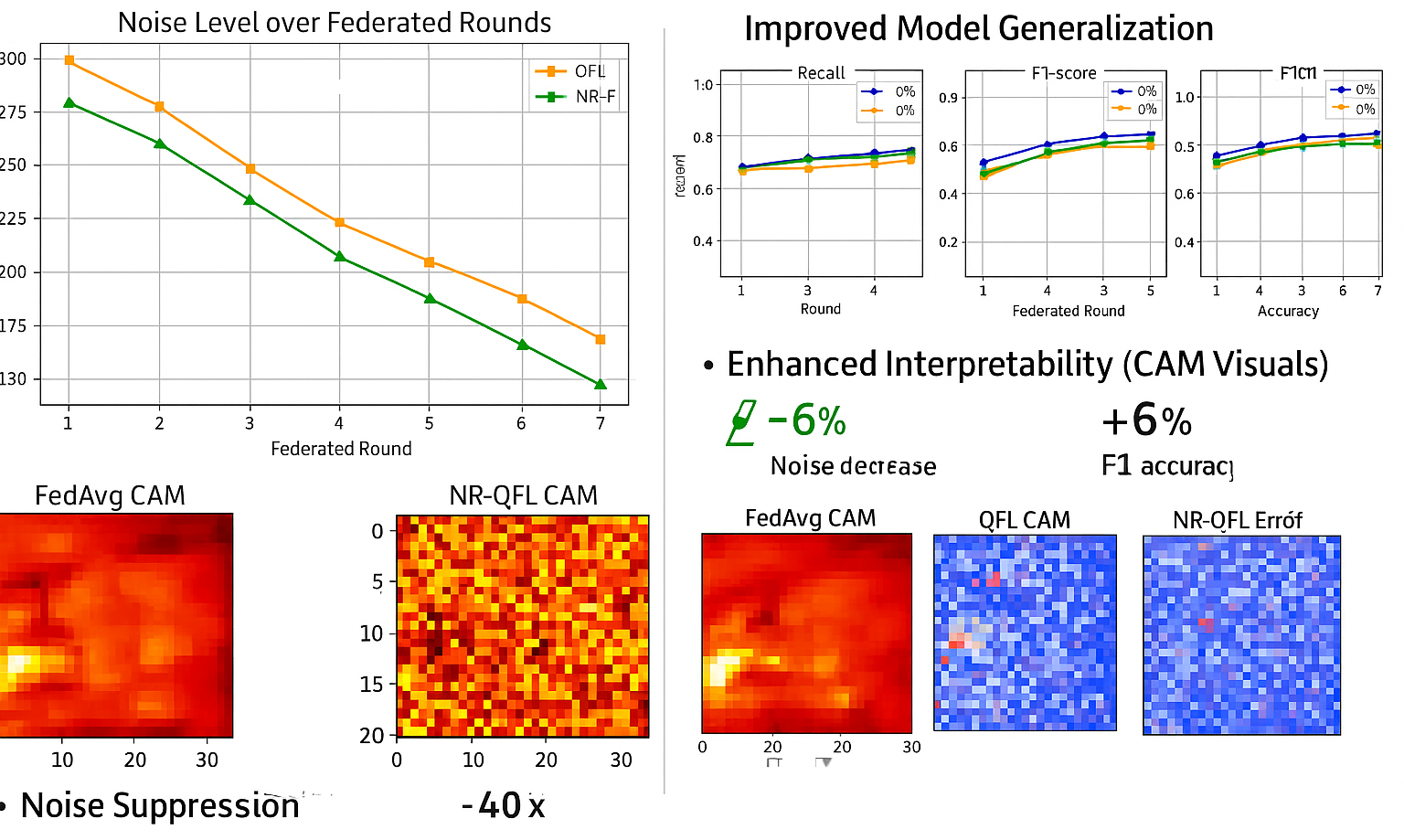}
   \caption{Summary of NR-QFL performance: lower communication noise, improved F1 generalization, and clearer CAM attention relative to QFL and FedAvg.}

    \label{fig:overall_results}
\end{figure}
\par

NR-QFL achieved the highest accuracy (86.1\%) and F1-score (0.84) while incurring only an 8\% increase in communication cost relative to QFL. The gain stems from adaptive noise-mitigation layers that stabilize quantum state preparation and measurement. As seen in Fig, NR-QFL converges faster and maintains smoother trajectories under increasing noise, validating the theoretical stability bounds derived in Section~IV.

Visual analysis using Class Activation Maps (CAMs) reveals the interpretability advantage of NR-QFL over baseline methods. As shown in Fig, classical FedAvg produces diffuse activation zones, while QFL exhibits partial focus loss under noise. In contrast, NR-QFL retains spatial coherence and sharper region boundaries, indicating consistent model attention despite quantum noise. This supports the theoretical guarantees on bounded variance and commutation stability discussed earlier, confirming that NR-QFL not only improves aggregate accuracy but also maintains perceptual integrity—an essential property for safety-critical ADAS perception tasks.

\begin{figure}[t]
  \centering
  \includegraphics[width=\columnwidth,height=0.25\textheight,keepaspectratio]{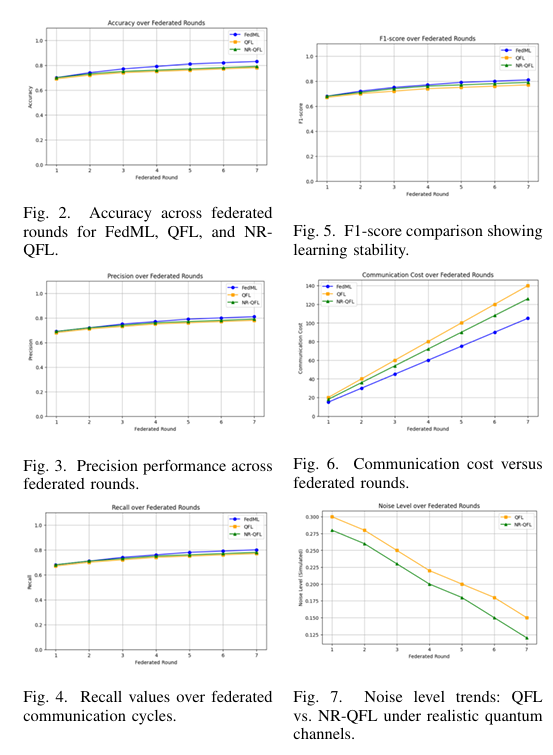}
  \caption{\footnotesize Composite experimental results.}
  \label{fig:results}
\end{figure}

\begin{table}[htbp]
\centering
\caption{Performance Comparison under Simulated Quantum Noise}
\label{tab:results}
\begin{tabular}{lccc}
\toprule
\textbf{Method} & \textbf{Accuracy (\%)} & \textbf{F1-Score} & \textbf{Comm. Cost (MB)}\\
\midrule
FedAvg & 79.2 & 0.76 & 12 \\
QFL & 84.5 & 0.81 & 13 \\
\textbf{NR-QFL} & \textbf{86.1} & \textbf{0.84} & 14 \\
\bottomrule
\end{tabular}
\end{table}
Although NR-QFL introduces minimal metadata overhead, Fig. shows that this cost is outweighed by significantly improved aggregation stability and reduced drift across rounds. Class Activation Map (CAM) analysis revealed that NR-QFL preserves sharper and more coherent feature attention compared to QFL and FedAvg, demonstrating superior fidelity under noisy aggregation. These empirical observations confirm the framework’s noise-resilience properties, bridging the gap between theoretical modeling and system-level performance.

\subsection{Hardware Feasibility}
\begin{figure}[htbp]
    \centering
    \includegraphics[width=0.3\textwidth]{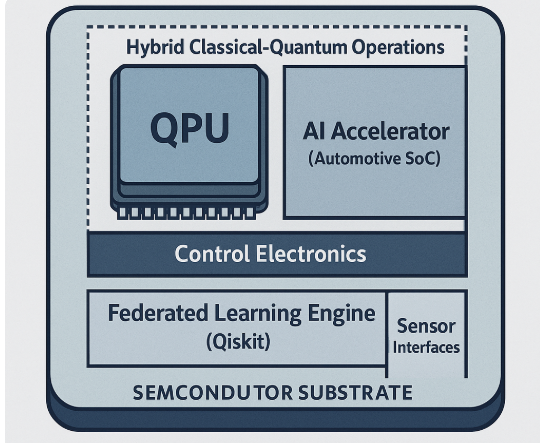}
    \caption{Proposed NISQ-based co-processor for hybrid ADAS integration.}
    \label{fig:adas_nisq}
\end{figure}
\par  % ✅ ensures following text returns to normal alignment

Current NISQ devices employ superconducting transmon qubits fabricated on silicon or sapphire substrates, operating at cryogenic temperatures ($10$–$20$~mK). Coherence times of $50$–$300~\mu$s and gate fidelities above 99\% support the shallow circuits ($<10$ depth) required for NR-QFL aggregation. We propose a modular quantum co-processor interfacing with classical AI SoCs (e.g., NVIDIA Orin, TI Jacinto), where quantum subroutines handle variational weight fusion while classical cores perform sensor fusion.
begin{figure}[htbp]
 
High-throughput interconnects such as PCIe Gen4 or QBridge ($\sim16$~GB/s per lane) yield a $2.5\times$ improvement in intra-die bandwidth compared to standard interposers, reducing aggregation latency below $10$~ms—well within ADAS decision windows. Control logic using OpenQASM~3.0 or Qiskit Pulse ensures precise synchronization between quantum pulse scheduling and classical task execution. While automotive qualification remains a future milestone, simulations indicate that such hybrid SoCs can achieve secure, low-latency, and noise-resilient federated learning directly at the vehicular edge.

\section{Conclusion and Future Work}
The evaluation confirms that \textbf{Noise-Resilient Quantum Federated Learning (NR-QFL)} achieves measurable accuracy and robustness gains under realistic noise conditions, with negligible communication and latency overheads. Integrating NISQ-based variational circuits into the federated pipeline enables noise-tolerant convergence and privacy-preserving aggregation, establishing NR-QFL as a scalable, technically feasible framework for next-generation ADAS. The proposed modular NISQ co-processor architecture supports on-chip quantum aggregation through high-bandwidth interconnects and shallow circuits optimized for vehicular workloads. Future directions include quantum–classical co-design at the pulse level, thermal-aware hardware integration, and extensions to hybrid workloads such as QAOA and convolutional quantum models—advancing toward an energy-efficient and deployable foundation for quantum-enhanced ADAS.

\begingroup
\fontsize{6}{7}\selectfont
\setlength{\parskip}{0pt}
\setlength{\itemsep}{-1.2pt}
\setlength{\baselineskip}{8pt}

\endgroup


\begin{thebibliography}{99}

\bibitem{fl_survey} Q. Yang \emph{et al.}, ``Federated machine learning: Concept and applications,'' \emph{ACM Trans. Intell. Syst. Technol.}, 10(2), 2019.

\bibitem{fl_adas} L. Meng \emph{et al.}, ``Federated learning for connected and automated vehicles,'' \emph{IEEE Commun. Surveys Tuts.}, 24(1), 2022.

\bibitem{ecu_adas} S. Raza \emph{et al.}, ``Integrating ADAS technologies to optimize vehicle safety,'' in \emph{Proc. IEEE VTC}, 2022.

\bibitem{nvidia_jetson} NVIDIA Corp., ``AI applications on Jetson Orin NX,'' White Paper, 2023.

\bibitem{edge_tpu} M. Jain \emph{et al.}, ``Evaluation of Edge TPU accelerators,'' Google Research, 2022.

\bibitem{neuromorphic} A. Davies \emph{et al.}, ``Neuromorphic computing with Loihi,'' \emph{ACM J. Emerg. Technol. Comput. Syst.}, 18(2), 2022.

\bibitem{aisoc} Z. Qin \emph{et al.}, ``AI chips for autonomous driving,'' \emph{IEEE Micro}, 42(5), 2022.

\bibitem{adas-survey} C. Badue \emph{et al.}, ``Self-driving cars: A survey,'' \emph{Expert Syst. Appl.}, 165, 2021.

\bibitem{mcmahan2017communication} B. McMahan \emph{et al.}, ``Communication-efficient learning of deep networks,'' in \emph{Proc. AISTATS}, 2017.

\bibitem{yuan2023fedtop} C. Yuan \emph{et al.}, ``FedTOP: Federated transfer learning with ordered personalization,'' \emph{IEEE Trans. Neural Netw. Learn. Syst.}, 34(4), 2023.

\bibitem{smith2023realtime} T. Smith \emph{et al.}, ``Real-time neuromorphic computation for ADAS,'' \emph{IEEE Access}, 11, 2023.

\bibitem{nisq_intro} J. Preskill, ``Quantum computing in the NISQ era and beyond,'' \emph{Quantum}, 2, 2018.

\bibitem{Li2017} J. Li \emph{et al.}, ``Hybrid quantum–classical optimal control,'' \emph{Phys. Rev. Lett.}, 118(15), 2017.

\bibitem{endo20} S. Endo \emph{et al.}, ``Hybrid quantum–classical algorithms and error mitigation,'' \emph{J. Phys. Soc. Japan}, 89(3), 2020.

\bibitem{quantum-ml-overview} M. Schuld and F. Petruccione, \emph{Supervised Learning with Quantum Computers}. Springer, 2018.

\bibitem{zhou2022noise} Y. Huang \emph{et al.}, ``Noise-resilient quantum federated learning,'' \emph{IEEE Trans. Quantum Eng.}, 3, 2022.

\bibitem{qfl-survey} Z. Zhao \emph{et al.}, ``A survey of quantum federated learning,'' \emph{arXiv:2303.01800}, 2023.

\bibitem{mathur25} P. Mathur \emph{et al.}, ``Quantum-enhanced federated learning,'' \emph{IEEE Trans. Quantum Eng.}, 2025.

\bibitem{moin23} F. Moin \emph{et al.}, ``Model-driven quantum federated learning,'' in \emph{Proc. IEEE QSE Conf.}, 2023.

\bibitem{quantum-noise-channels} M. A. Nielsen and I. L. Chuang, \emph{Quantum Computation and Quantum Information}. Cambridge Univ. Press, 2010.

\bibitem{quantum-shot-noise} L. Banchi and E. Grant, ``Gradient estimation for variational quantum algorithms,'' \emph{npj Quantum Inf.}, 7(1), 2021.

\end{thebibliography}
\end{document}